# Clustering Techniques for Marbles Classification

*Caldas Pinto, J.R\*., Pina, P.\*\*, Ramos V. \*\*, Ramalho M.\**

*IDMEC/IST - Technical University of Lisbon - Instituto Superior Técnico,
\*\*CVRM/Centro de Geo-Sistemas - Instituto Superior Técnico
Av. Rovisco Pais, 1049-001 Lisboa, PORTUGAL
{jcpinto,mar }@dem.ist.ist.utl.pt,{ppina,vitorino.ramos}@alfa.ist.utl.pt

*Abstract* – **Automatic marbles classification based on their visual appearance is an important industrial issue. However, there is no definitive solution to the problem mainly due to the presence of randomly distributed high number of different colours and its subjective evaluation by the human expert. In this paper we present a study of segmentation techniques, we evaluate they overall performance using a training set and standard quality measures and finally we apply different clustering techniques to automatically classify the marbles.**

## I. INTRODUCTION

Ornamental stones are quantitatively characterised in many ways, mostly physical, namely geological-petrographical and mineralogical composition, or by mechanical strength. However, the properties of such products differ not only in terms of type but also in terms of origin, and their variability can also be significant within a same deposit or quarry. Though useful, these methods do not fully solve the problem of classifying a product whose end-use makes appearance so critically important. Appearance is conditioned not only by the kind of stone but it also depends on the subjective evaluation of beauty. Traditionally, the selection process is based on human visual inspection, giving a subjective characterisation of the appearance of the materials instead of an objective reliable measurement of the visual properties (colour, texture, shape and dimensions of their components) and the identification of tolerances with respect to the accepted standards. Yet, quality control is essential in order to keep marble industries competitive: shipments of finished products (e.g. slabs) must be of uniform quality, and the price demanded for a product, particularly a new one, must somehow be justified. Thus, it is very important to have a tool for the objective characterisation of appearance.

In this work we establish a set of suitable tools to characterise the appearance of natural stones, supported by digital image analysis. In section II we present a brief description of the techniques that we use for marbles segmentation, one of them based on colour homogeneity reasoning and the other one on mathematical morphology. The quality of the descriptors is compared using quality measures. In section III we present two clustering techniques used in this project, one supervised – Learning Vector Quantisation (LVQ) and the other one unsupervised - Simulated Annealing, (SA). .Finally, we present results that will demonstrate the validity of the approaches developed in this work.

## II. MARBLES SEGMENTATION TECHNIQUES

### A. Based on Colour Homogeneity Principles

In a recent paper we presented a clustering algorithm for colour segmentation [1], [2]. This algorithm corresponds to a quadtree based colour homogeneity analysis. We reach the concept of colour homogeneity by experimentally analysing histograms of the RGB components of images that we perceive as homogeneous. Our goal was to find all the regions that could be considered to have visually the same colour. This was achieved using a quadtree approach. For each region in the quadtree division of the image, these histograms were tested under the hypothesis of corresponding to a Gaussian distribution. If the hypothesis does not fail a homogeneity condition is verified and characterised by the mean of the distribution.

This process of image division will stop if one of the following conditions is reached:
- All the regions are homogeneous
- The dimension of a given non-homogeneous region has an area lower than a given threshold.

The results of this algorithm are illustrated on fig. 1.

Together with the original image we present the so-called marble signature that represents all set of regions obtained in the quadtree analysis. This signature suggest us that we can extract a set of meaningful features related to these regions. Before going on with their description it is necessary to advance one more step concerning our original goal of colour clustering.

Because our goal was to get a reduced number of colour regions is now necessary to attribute the same colour to regions that are not connected but indeed can not be visually distinguished. This is an important problem in image coding [3] and image retrieval, [4]. Then our algorithm follows again our way of perception. Indeed we are subjectively attracted by the bigger areas when questioned about a given region colour. This suggests the following methodology:
- To order the different encountered regions according to their area.
- The first region in the rank is compared with the remaining ones. Those regions that match a given distance criterion will have their colour feature replaced by the colour feature of the first region. These regions are taken out of the list.

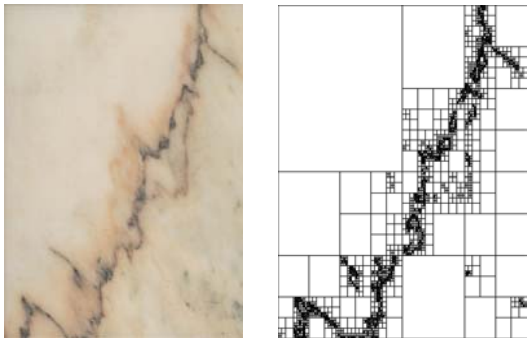

Fig. 1. (a) marble (b) marble signature showing the homogeneous regions obtained by our quadtree approach.

To measure visual distinction between colours we used one of the basic distance measures – Manhattan or Euclidian. This distances where measured in the HSV space, generally more adequate to this purpose.

As we can see in [1,2] this method leads to quite successful solutions.

All this process of image decomposition and reconstruction suggest us to keep some of the information involved in the process. In this study the parameters used as future image features for each final colour cluster are:
- Total area
- Number of connected regions that originate that area
- The R G B values attributed to the area (RGB of the largest connected region)
- The average and variance of the HSV values of the connected regions.

It can also be used, as a feature, the number of connected regions not classified in one of the five main colour clusters. Concluding, we obtained eleven features for each colour region, hence, a total number of fifty six features for each stone, considering five colours regions.

*B. Based on Mathematical Morphology*

Generically, the extraction of features by means of mathematical morphology operators is implemented at a global analysis stage. The main idea at this phase, which is also its main expectation, is that the extracted features are sufficient to discriminate several types of samples avoiding local analysis operations. Features, that characterise the different structures, concern the colour, the size/shape and the texture of their components.

**Colour features (54)**

Colour features are based on the grey level distributions of each channel in the colour systems Red-Green-Blue (RGB) and Hue-Intensity-Saturation (HSV). In order to reduce the amount of information, octils of each curve were considered instead of considering the frequency for all the 256 grey levels. Besides these 8 values, the minimum value of each distribution was also considered. Thus, in what concern colour features, 54 values were considered: R(8+1), G(8+1), B(8+1), H(8+1), S(8+1), V(8+1).

**Size/shape and textural features (540)**

The size distribution by grey level granulometry can be computed by morphological openings or closings of increasing size. Both operators have granulometric properties [5] because they are increasing, extensive (closing) and anti-extensive (opening) and idempotent [6]. It consist on measuring the volume $V$ of the function $f$ after the application of the opening, $\gamma^{B(r)}(f)$, or closing, $\varphi^{B(r)}(f)$, with the structuring element $B$ of size $r$. The granulometry by opening reflects the dimension of the lighter components (peaks) while the granulometry by closing informs about the dimension of darker components (valleys). But, besides the information related to the size/shape given by these classical grey level granulometries (measure of the volume), another measure can be achieved by associating to it the distribution of the grey levels. This measure, recently introduced [7], consists on the computation of the grey level distribution for opened or closed images of increasing size. The resulting diagram (named size-intensity) is bidimensional and incorporates this way both size/shape and intensity information. This diagram is built using a family of openings or closings of increasing size using the cylindrical structuring element $B(r,k)$ of radius $r$ and height $k$.

For each sample, 30 granulometrical classes were considered, corresponding each one to the application of an opening and a closing with a hexagonal structuring element. Similarly to the colour features, the octils of each granulometrical class and its minimum were considered. This way we have 540 features: (8+1)*30 for the openings plus (8+1)*30 for the closings.

**Mean interset distance**

The mean interset distance is the mean distance (in the feature space) between two objects from different clusters. The bigger the interset distance, the better the cluster set is. The interset distance for two given clusters $c1$ and $c2$ is:

$$D_{\text{inter}}^{c1,c2} = \frac{1}{M_1 M_2} \times \sum_{i}^{M_1} \sum_{j}^{M_2} d^2(x_i, x_j)$$

where $d^2(x_i, x_j)$ is the square distance between two objects in the feature space, and $M_1$ and $M_2$ are the number of objects is clusters $c1$ and $c2$, respectively. The mean interset distance is:

$$\overline{D_{\text{inter}}} = \frac{1}{N_c(N_c - 1)} \times \sum_{c1} \sum_{c2 \neq c1} D_{\text{inter}}^{c1,c2}$$

where $N_c$ is the number of clusters in the cluster set.

*Intraset/Interset combined distance*

Combining both previous quality measures, we obtain the intraset distance divided by the interset distance. The lower the intraset/interset, the better the cluster set is. The **interset/interset combined distance** is:

$$D_{\text{intra/inter}} = \frac{\sum_{c} D_{\text{intra}}^{c}}{\sum_{c1} \sum_{c2 \neq c1} D_{\text{inter}}^{c1,c2}}$$

These quality measures were used to evaluate how the used features are adequate to separate the clusters defined by experts.

### III. CLUSTERING TECHNIQUES

To evaluate the features adequacy, two clustering techniques were used and results compared. It was chosen a supervised technique – Learning Vector Quantization (LVQ), [9] and an unsupervised one – Simulated Annealing (SA), [10].

### IV. RESULTS

The presented algorithms will now be applied to a set of Portuguese marbles. The main goal of this work is to evaluate our derived set of features for two different purposes: dominant colour clustering and vein density clustering. Because, in practice, the division according to the vein density comes after the base colour classification, we will do our tests with clustering according to the veins only for one chosen colour.

We evaluate and present the results in two different ways. In the first one we organise a table with some parameters concerning the mean intraset and interset distance and the ratio between them. It is also useful in the analysis of the results to present also the maximum intraset distance for the set of clusters and the minimum distance concerning all the pairs of clusters. It is obvious if that maximum is lower that this minimum the classification may give some wrong results.

The above discussion will lead us to a set of more promising features for clustering. For this set, we will apply the LVQ clustering technique to find the prototypes of the different clusters and after that to classify the same training set. We also use the SA algorithm to blindly divide our training set in the desired number of clusters. Results will be present though the confusion matrix.

Naturally the best way to evaluate the results would be to present the images of the marbles classified in each cluster. They will be presented in an annex of this paper (only in its CD version).

In order to understand the tables is necessary to introduce some conventions.

The marbles were classified in 6 classes of colours and three classes of vein density (almost no veins, light veins and strong veins).

There are 56 features based on colour homogeneity (HF) that corresponds to:

- 1+5k to 5+5k ( k=0,4): Respectively area, number of fragments, R,G, B of the largest connected area of the k colour, mean and standard deviation of H,S,V of all the fragments that were merged in colour k.
- Feature 56: number of fragments not included in the 5 colour clusters.

Because is out of question to analyse all possible combinations of these filters we selected three paradigmatic groups:
A: all the 56 features
B: RGB components and mean HSV components for the colour one (that has bigger area)
C: (related to veins)

There are 594 features based on morphological analysis (MF) originated by:
- They are defined three histograms respectively for the R,G and B components. For each histogram:
- feature 1: min. grey value
- feature 2 to 8 : Grey Value - $r^{st}$ Octil
- 9 * 30 ($\gamma^1(f), \dots \gamma^{30}(f)$)
- 9 * 30 ($\varphi^1(f), \dots \varphi^{30}(f)$)

For the reason given above we chose three subsets of these clusters:
- A: all features (594).
- B: Features 1 to 324 (related to colour)
- C: Features 325 to 594 (related to the veins)

Finally we note that original data was normalised to be zero mean with unit variance.

Tables I to IV summarise the results of the experiments carried out with the described features and algorithms. In the digital version of this paper we include, as illustrative purpose, pictures from the Colour Clustering using the SA algorithm and the Vein Clustering using the LVQ.

| Table I: Quality Measures for Colour and Vein Clustering | | | | | | |
|---|---|---|---|---|---|---|
| | Colour Clustering | | | Vein Clustering | | |
| | Intra | Inter | Comb. | Intra | Inter | Comb |
| HF_A | 85.9 | 110.0 | 0.781 | 129.1 | 120.9 | 1.068 |
| HF_B | 2.32 | 13.34 | 0.174 | 1.78 | 2.79 | 0.637 |
| HF_C | 20.9 | 23.6 | 0.886 | 9.58 | 27.4 | 0.349 |
| MF_A | 255.9 | 1278.7 | 0.200 | 187.2 | 370.8 | 0.505 |
| MF_B | 133.4 | 685.5 | 0.195 | 122.4 | 216.0 | 1.130 |
| MF_C | 122.4 | 593.2 | 0.206 | 54.3 | 154.8 | 0.351 |

| Table II: confusion matrix for LVQ (colour clust.) | | | | | | | | | | | |
|---|---|---|---|---|---|---|---|---|---|---|---|
| | Features HF_B | | | | | | Features MF_B | | | | |
| 1 | 8 | 1 | 0 | 0 | 0 | 0 | 8 | 1 | 0 | 0 | 0 | 0 |
| 2 | 0 | 8 | 0 | 0 | 0 | 0 | 0 | 8 | 0 | 0 | 0 | 0 |
| 3 | 0 | 0 | 7 | 0 | 0 | 0 | 0 | 4 | 3 | 0 | 0 | 0 |
| 4 | 0 | 0 | 2 | 8 | 0 | 0 | 0 | 0 | 0 | 10 | 0 | 0 |
| 5 | 0 | 0 | 0 | 4 | 16 | 9 | 0 | 0 | 1 | 6 | 11 | 11 |
| 6 | 0 | 0 | 0 | 0 | 0 | 6 | 0 | 1 | 0 | 2 | 0 | 3 |

| Table III: confusion matrix for SA (colour clust.) | | | | | | | | | | | |
|---|---|---|---|---|---|---|---|---|---|---|---|
| | Features HF_B | | | | | | Features MF_B | | | | |
| 1 | 6 | 3 | 0 | 0 | 0 | 0 | 5 | 4 | 0 | 0 | 0 | 0 |
| 2 | 0 | 6 | 2 | 0 | 0 | 0 | 0 | 6 | 2 | 0 | 0 | 0 |
| 3 | 0 | 0 | 7 | 0 | 0 | 0 | 0 | 0 | 7 | 0 | 0 | 0 |
| 4 | 0 | 0 | 6 | 1 | 3 | 0 | 0 | 0 | 0 | 9 | 0 | 1 |
| 5 | 0 | 0 | 1 | 12 | 13 | 3 | 0 | 0 | 2 | 6 | 11 | 10 |
| 6 | 0 | 0 | 0 | 0 | 3 | 3 | 0 | 0 | 1 | 3 | 1 | 1 |

| Table IV: confusion matrix for LVQ/SA (vein clust.) | | | | | | | | | | | |
|---|---|---|---|---|---|---|---|---|---|---|---|
| | LVQ/HF_C | | | LVQ/MF_C | | | SA/HF_C | | | SA/MF_C | | |
| 1 | 7 | 2 | 0 | 5 | 4 | 0 | 7 | 0 | 2 | 6 | 3 | 0 |
| 2 | 3 | 4 | 0 | 0 | 6 | 1 | 5 | 0 | 2 | 0 | 6 | 1 |
| 3 | 0 | 1 | 12 | 0 | 0 | 13 | 0 | 4 | 9 | 0 | 1 | 12 |

Table I shows that a correct choice of the features according to the type of desired clustering considerably improves the quality measures. In this table features referred as B were mainly appropriate to colour characterisation and those referred as C to vein characterisation. Considering all the features (A) leads always to worst results.

Tables II to IV summarises two different experiments, a supervised clustering technique (LVQ) and an unsupervised one (SA). Because our goal was to evaluate our features, the training set and the test set used with the LVQ was the same. The same data was used with the SA. Two sets of data were considered: one for colour (69 samples) and one for vein (29 marbles with similar background colour). The results show the confusion matrixes. In general we get a satisfactory degree of matching, despite some difficulties with certain colour classes. We should emphasise that to elaborate a training set is not a straightforward process due to the richness and variety of these natural textures, that causes that any classification by experts has always a non negligenciable degree of subjectivity.

## V. CONCLUSIONS

Two main conclusions can be addressed with this paper. Firstly, both of the two approaches used to extract features of marbles are adequate to cover all the visual characteristics of the marbles. Anyhow it is advisable to make a previous selection of the respective descriptors in order to reduce its initial number. Secondly, the two clustering techniques used have shown to be adequate since have led to satisfactory classification rates. However, these results can still be improved, since there was no effort to optimise the used clustering algorithms. Many others algorithms could be used like those based on neural networks or fuzzy clustering. Further work will concern them.

## ACKNOWLEDGEMENTS

This project is supported by the Portuguese National Project CVAM – PRAXIS XXI/2/2.1/TPAR/2057/95 and by program POCTI, FCT, Ministério da Ciência e da Tecnologia, Portugal.

**Images from the clusters described in Table III**

**Results using the SA algortithm for clustering**

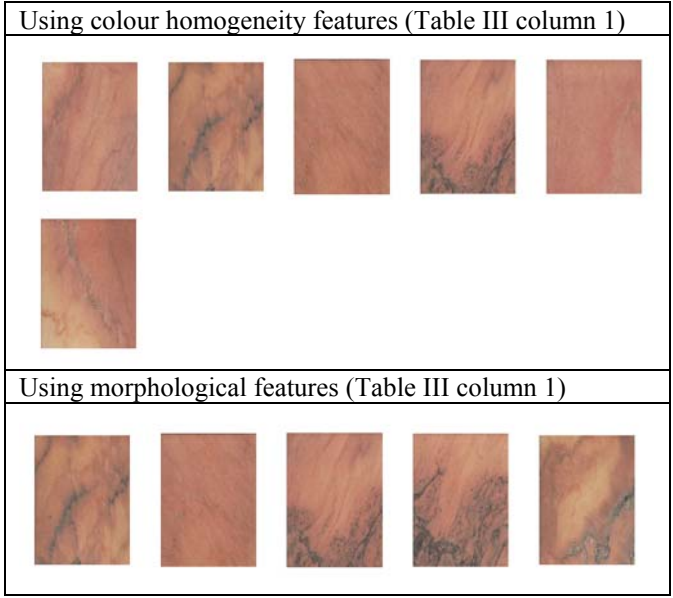

Using colour homogeneity features (Table III column 1)

Using morphological features (Table III column 1)

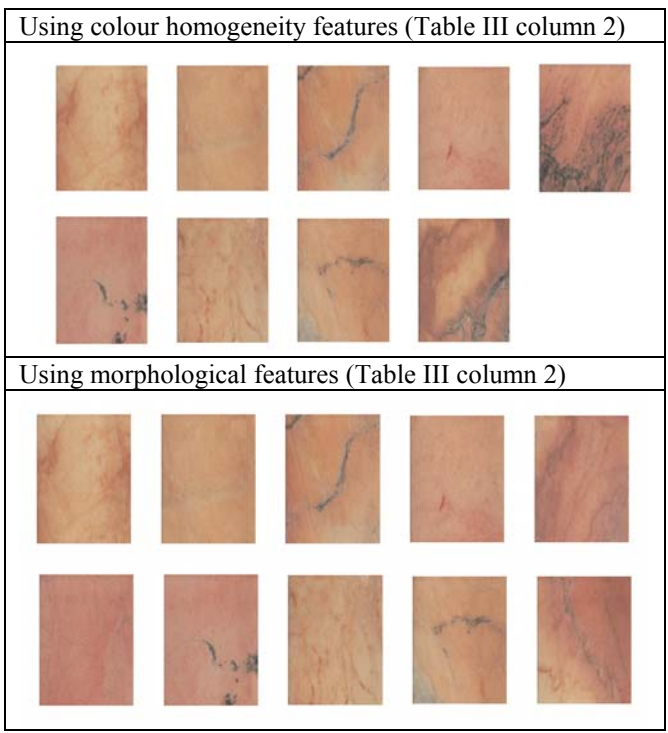

Using colour homogeneity features (Table III column 2)

Using morphological features (Table III column 2)

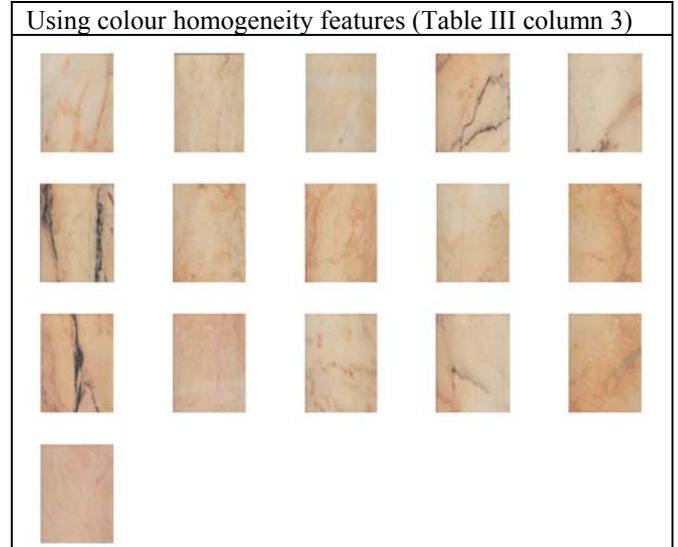

Using colour homogeneity features (Table III column 3)

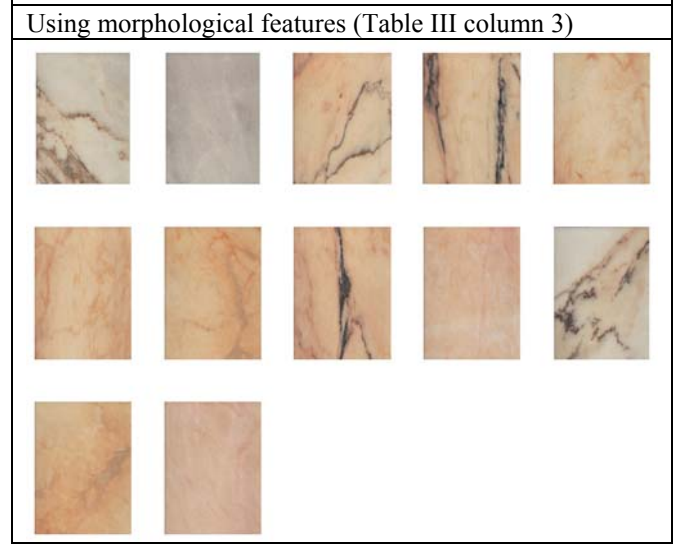

Using morphological features (Table III column 3)

Using colour homogeneity features (Table III column 4)

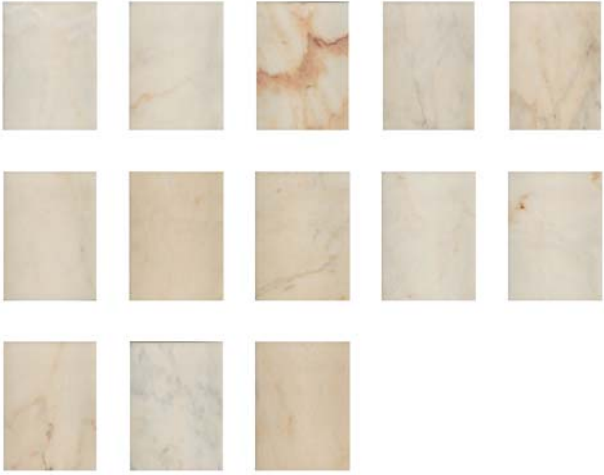

Using colour homogeneity features (Table III column 5)

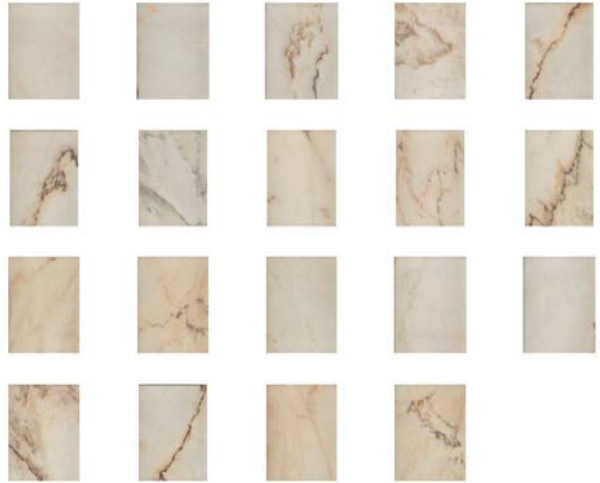

Using morphological features (Table III column 4)

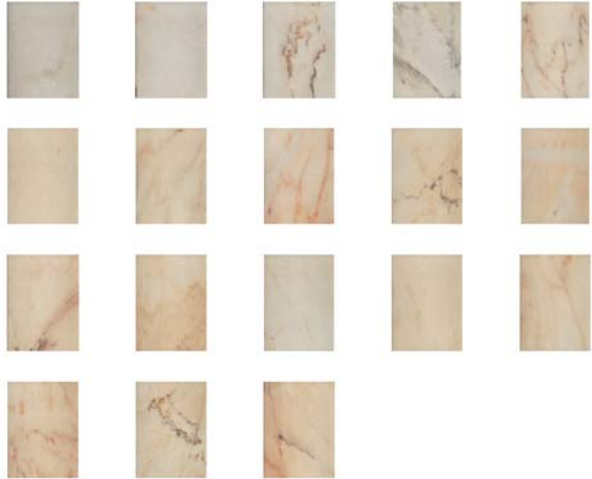

Using morphological features (Table III column 5)

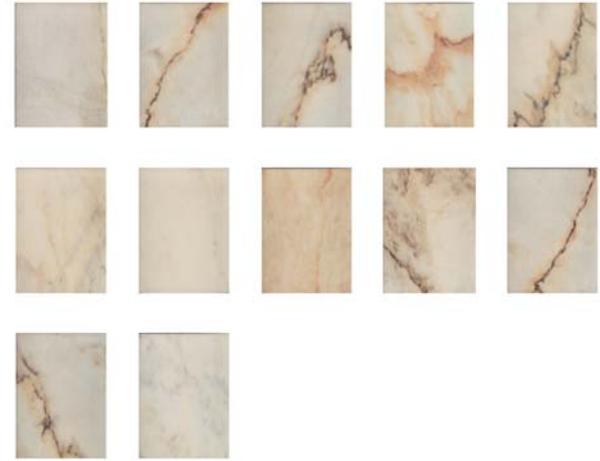

Using colour homogeneity features (Table III column 6)

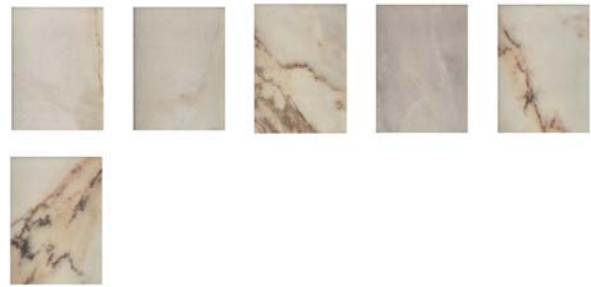

Using morphological features (Table III column 6)

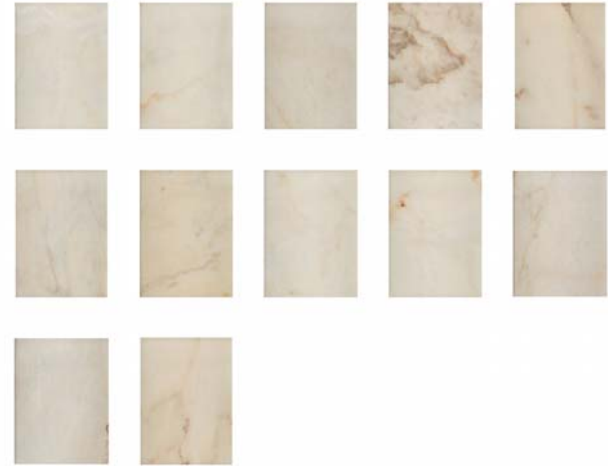

**Images from the clusters described in Table IV**

**Results using the LVQ algortithm for clustering**

| **Results using the LVQ algortithm for clustering Class 1 (Table IV)** |
|---|
| Using colour homogeneity features |
| 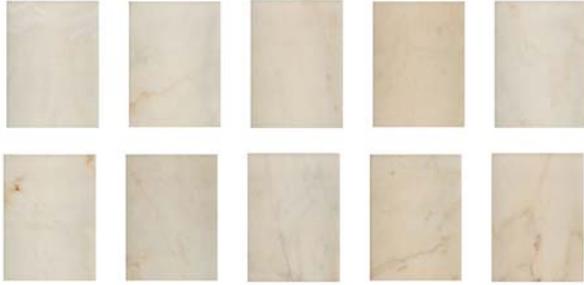 |
| Using morphological features |
| 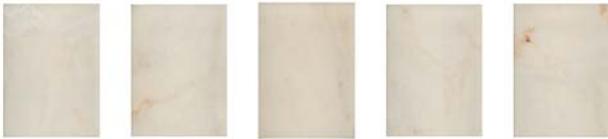 |

| **Results using the LVQ algortithm for clustering Class 3 (Table IV)** |
|---|
| Using colour homogeneity features |
| 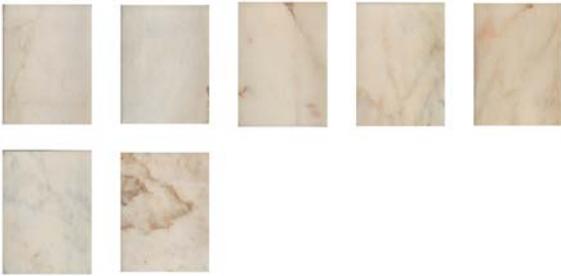 |
| Using morphological features |
| 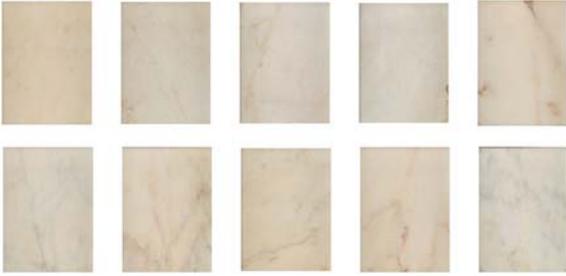 |

| **Results using the LVQ algortithm for clustering Class 2 (Table IV)** |
|---|
| Using colour homogeneity features |
| 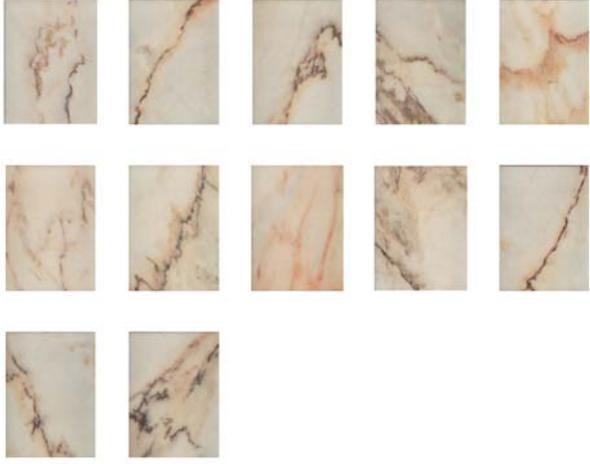 |
| Using morphological features |
| 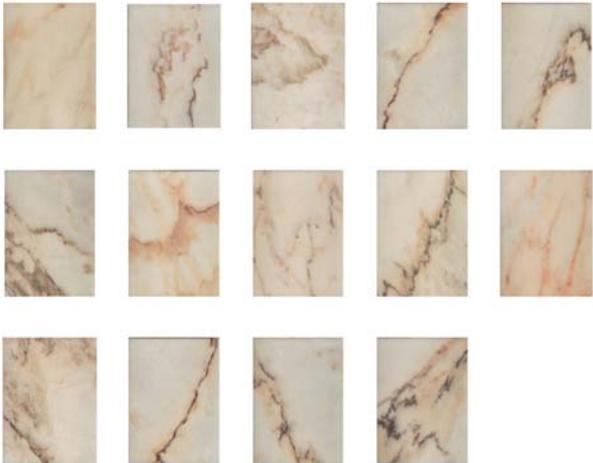 |